\documentclass[10pt,twocolumn,letterpaper]{article}

\usepackage[pagenumbers]{cvpr} 
\usepackage[pagebackref,breaklinks,colorlinks,allcolors=cvprblue]{hyperref}

\usepackage{multirow,tabularx}
\usepackage{marvosym}

\definecolor{cvprblue}{rgb}{0.21,0.49,0.74}

\title{TempoMaster: Efficient Long Video Generation via Next-Frame-Rate Prediction}

\author{
Yukuo Ma$^{1,2}$
\and
Cong Liu$^{2}$
\and
Junke Wang$^{1}$
\and 
Junqi Liu$^{2}$
\and 
Haibin Huang$^{2}$
\and 
Zuxuan Wu\Letter$^{1}$
\vspace{0.01cm}
\and 
Chi Zhang$^{2}$
\and 
Xuelong Li\Letter$^{2}$
\vspace{0.05cm}
\and 
$^1$Fudan University \hspace{2em}
$^2$Institute of Artificial Intelligence (TeleAI), China Telecom
\vspace{0.01cm}
\and
{\tt\small ykma25@m.fudan.edu.cn}
\and
{\tt \small zxwu@fudan.edu.cn}
\and
{\tt\small xuelong\_li@ieee.org}
}

\begin{document}

\twocolumn[{
\maketitle
\begin{center}
\includegraphics[width=\linewidth]{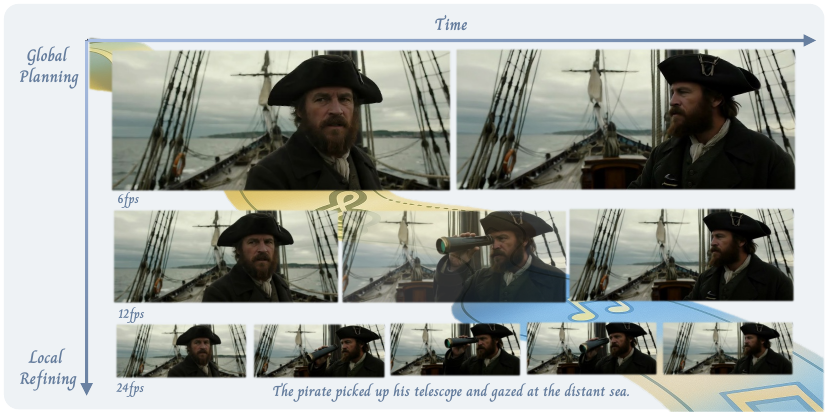}
\end{center}
\vspace{-0.5cm}
\captionsetup{type=figure}
\captionof{figure}{
\textbf{TempoMaster} first generates a video sequence at coarse and low frame rate to establish the global dynamics and semantic structure, and subsequently refines it by predicting frames at higher rates, thereby enhancing temporal smoothness and detail.
This next-frame-rate prediction paradigm results in videos with improved motion quality and temporal consistency.
}\label{fig:teaser}
\vspace{0.3cm}
}]

\begin{abstract}
We present TempoMaster, a novel framework that formulates long video generation as next-frame-rate prediction. Specifically, we first generate a low-frame-rate clip that serves as a coarse blueprint of the entire video sequence, and then progressively increase the frame rate to refine visual details and motion continuity. During generation, TempoMaster employs bidirectional attention within each frame-rate level while performing autoregression across frame rates, thus achieving long-range temporal coherence while enabling efficient and parallel synthesis. 
Extensive experiments demonstrate that TempoMaster establishes a new state-of-the-art in long video generation, excelling in both visual and temporal quality. See our project page at \href{https://scottykma.github.io/tempomaster-gitpage/}{https://scottykma.github.io/tempomaster-gitpage/}
\end{abstract}
    
\begin{figure*}
    \centering
    \includegraphics[width=\textwidth]{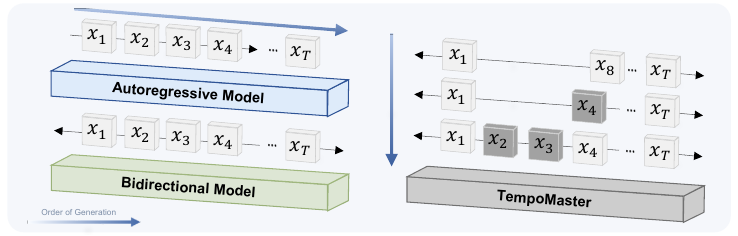}
    \caption{
    \textbf{Different video modeling paradigms.}
    Autoregressive models generate frames sequentially under a causal structure. Bidirectional models generate the entire sequence at once by processing the full sequence directly. TempoMaster establishes the global structure via a low-frame-rate bidirectional pass, then progressively enhances local details via predicting the video at the next higher frame rate.
    }
    \label{fig:architecture}
\end{figure*}

\section{Introduction}
\label{sec:intro}

Video generation with deep generative models has advanced rapidly in recent years, with remarkable improvements in both fidelity and controllability. These advances have enabled practical applications in film production~\cite{moviedreamer2024,moviegen2024,captain2025}, interactive storytelling~\cite{yan2025,longlive2025}, neural game engines~\cite{hunyuan-gamecraft2025}, and world modeling~\cite{genie2024,genie3}. Despite the progress, generating videos that are visually natural and temporally coherent over long horizons remains challenging~\cite{exposurebias2024}, due to the dual challenges of maintaining long-term consistency and incurring exorbitant computational expense.

Current approaches for long video generation fall into two primary paradigms. The first treats an entire video as a single spatiotemporal volume and applies bidirectional attention across all frames~\cite{wan2025,waver2025,sora2024,seedance2025}. This design excels at modeling long-range temporal coherence but incurs quadratic growth in memory and computation, making long-video generation prohibitively expensive.

The other generates long video sequentially in an autoregressive manner~\cite{videogpt2021,cogvideo2023,longlive2025,lvd2022,causvid2025, videopoet2024, ttt2025}. 
Specifically, recent efforts~\cite{df2024,self-forcing2025,causvid2025,framepack2025,longlive2025,ttt2025} iteratively predict future video clips conditioned on previously generated segments to extend the temporal horizon.
Although this strategy improves the efficiency of long-video generation, it inevitably requires maintaining a continuously growing history of frames.
To avoid the prohibitive computational cost of processing an unbounded context, existing methods typically truncate~\cite{magi2025,rollingdiffusion2024,rolling-forcing2025} or compress past frames~\cite{framepack2025,pyramidflow2025}, generating only a short chunk at a time.
However, such constraints cause the model to forget earlier content, and the iterative generation process allows minor prediction errors to accumulate over time, gradually resulting in appearance drift and motion inconsistencies.

The complementary strengths and limitations of the above two methods motivate us to raise a question: can we integrate both to achieve a better balance between temporal consistency and inference efficiency? Considering the substantial temporal redundancy in videos, we posit that the key lies in decoupling the generation of high-level temporal semantics from low-level visual details.  In other words, a coherent dynamic structure can be established using only a sparse subset of keyframes, while the remaining intermediate frames can be efficiently inferred based on the learned temporal dynamics and contextual dependencies.

With this in mind, this paper presents \textbf{TempoMaster}, a novel framework that decouples long-term temporal structuring from local frame synthesis within a \textbf{next-frame-rate prediction }paradigm. As illustrated in Fig.~\ref{fig:architecture}, our model first predicts a sparse and low-frame-rate blueprint sequence that summarizes the global temporal dynamics. Based on this, we progressively increase the frame rate through multiple refinement stages to fill in fine-grained motion and appearance details. This hierarchical generation strategy gradually enriches temporal details while maintaining global semantic consistency. As a result, full-frame-rate videos can be produced efficiently without processing every frame simultaneously. Moreover, this formulation enables parallel generation across temporal segments, substantially reducing computational cost while preserving long-range temporal coherence.
In summary, the contributions of our work are outlined as follows:
\begin{itemize}
    \item We propose TempoMaster, a simple yet effective method that integrates the global planning capacity of bidirectional modeling with the progressive generation of autoregressive approaches to address temporal consistency in long video generation.
    \item We propose an efficient generation strategy that enables parallel generation of high-frame-rate segments across temporal domains, significantly reducing the computational complexity of long video generation.
    \item Extensive experimental results demonstrate that our method can flexibly generate high-quality videos ranging from a few seconds to dozens of seconds in length, while exhibiting superior temporal consistency.
\end{itemize}

\section{Related Work}

\subsection{Short Video Generation}

Current approaches for short video generation can be broadly categorized into two classes: diffusion-based models \cite{ddpm, alyl2023, lvd2022} and autoregressive-based models \cite{videogpt2021, cogvideo2023, magi2025, nova2025}. Diffusion-based methods typically employ U-Net \cite{unet} or DiT \cite{dit} architecture to synthesize videos with an iterative denoising process. Notably, several works \cite{sora2024, wan2025, seedance2025, waver2025} demonstrate that by scaling up Diffusion Transformer (DiT) models and pretraining on large-scale video datasets, the visual quality and temporal coherence of synthesized videos can be consistently improved.

Autoregressive models typically discretize video clips into sequences of visual tokens and train an autoregressive transformer model to predict them one-by-one \cite{videogpt2021, videopoet2024, cogvideo2023}. These approaches are limited by the inference efficiency, as the large number of tokens in video sequences brings significant computational expenses. Recent efforts have integrated diffusion and autoregressive processes to improve video generation quality. Some take initial frames as a condition and train a diffusion model to denoise the following frames \cite{extdm2024, pyramidflow2025, causvid2025}, while others denoise the entire video sequence but assign lower noise levels to earlier frames during training \cite{rollingdiffusion2024, df2024, hgvd2025, magi2025}. 

\subsection{Long Video Generation}
Beyond clip generation, long video generation faces challenges in both computational complexity and long-term temporal consistency. While powerful bidirectional generative models like DiT~\cite{peebles2023scalable} have achieved remarkable success in short videos, their computational costs increase quadratically with the sequence length, making it difficult to extend to long video generation directly. Autoregressive models inherently support rollout of video clips along the temporal dimension, but often struggle with maintaining spatiotemporal coherence over extended periods~\cite{cogvideo2023, videogpt2021, videopoet2024, nova2025}. Several approaches attempt to weaken the error propagation by degrading historical frames through re-noising or masking \cite{df2024, magi2025, hgvd2025, rollingdiffusion2024}, and using their own inference results as historical frames \cite{self-forcing2025, rolling-forcing2025, self-forcing++2025}. More recently, a series of works have employed anchor-frame generation models to first produce key frames corresponding to different temporal stages of a video based on textual descriptions, and then synthesize the intermediate content between these anchors~\cite{nuwaxl2023,videostudio2024,captain2025,moviedreamer2024,storyagent2024}. These approaches have achieved impressive results in long-video generation and multi-shot video synthesis. However, they introduce additional complexity during both training and inference, including the need for a separate reference-frame generator and specialized procedures for planning and generating anchor frames at inference time.
\section{Method}
\label{sec:method}

We introduce TempoMaster, a framework for generating temporally coherent videos under varying frame rates.
Section~\ref{method.1} reformulates long-range video generation to naturally support flexible temporal resolutions.
Section~\ref{method.2} then presents the architecture built upon this formulation to maintain stable long-term dynamics.
Section~\ref{method.3} describes our training strategy on multi–frame-rate data, enabling the model to learn a continuous temporal representation.
Finally, Section~\ref{method.4} outlines our parallel inference strategy for efficient long-video synthesis with preserved global temporal consistency.
\begin{figure}[t]
  \centering
   \includegraphics[width=\linewidth]{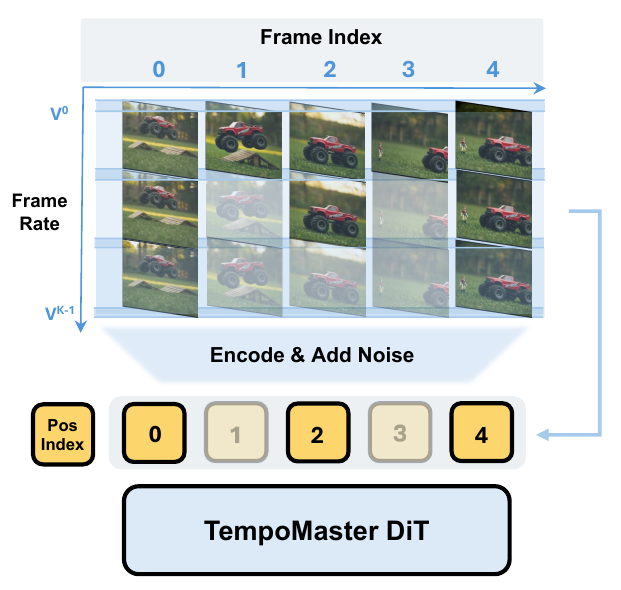}

   \caption{
   \textbf{Multi-Frame-Rate Training.} TempoMaster is trained on videos with varying frame rates, which are signaled to the model by scaling the interval of the temporal positional indices. As illustrated, training on a video at half the highest frame rate employs a positional index interval of 2.
   }
   \label{fig:3}
\end{figure}

\subsection{Next-Frame-Rate Prediction} \label{method.1}

Consider a sequence of frames $V=\left(x_1, x_2, \dots, x_T\right)$. As illustrated in Fig.~\ref{fig:architecture}, bidirectional models directly model the joint likelihood of all frames $p(V)$, while autoregressive models typically factorize it into:

\begin{equation}
p\left(V\right) = \prod_{t=1}^Tp(x_t|x_0, x_1,...,x_{t-1} )
\end{equation}

This formulation, known as next-frame prediction, shares the principle of sequentially generating conditioned on historical context with other autoregressive variants.      
In contrast, our method discards sequential historical conditioning, instead generating the frame sequence via next-frame-rate prediction.

We define a set of $K$ sequences at different frame rates, denoted by $V^0, V^1, \dots, V^{K-1}$. The $i$-th sequence $V^i = (x_{m}, x_{2m}, \dots, x_{jm})$ is temporally subsampled from the original video with a stride of $m=2^i$, resulting in a length of $j=T/m$. And we reformulate the likelihood of the video as:
\begin{equation}
p(V)=p(V^{K-1})\prod_{i=0}^{K-2}p(V^i|V^{i+1},V^{i+2},...,V^{K-2})    
\end{equation}

As depicted in Fig.~\ref{fig:3}, we train a Diffusion Transformer (DiT) on videos spanning $K$ frame rates. 
This allows TempoMaster to model longer and more dynamic sequences at lower frame rates while maintaining the fixed context length of short video models.
During inference, we start with predicting the lowest-frame-rate video $V^{K-1}$, which effectively structures the global temporal dynamic of the whole video at once. 
Each subsequent stage takes all the previously generated frames as anchors and focuses on refining the in-between motion details. 
As the global content is predetermined, the generated frames can be divided into multiple short chunks and fed to the model in parallel. 
In contrast to truncating and compressing, our parallel inference strategy could deal with the growing length of frames without loss of generation quality.

\begin{figure}[t]
  \centering
   \includegraphics[width=0.9\linewidth]{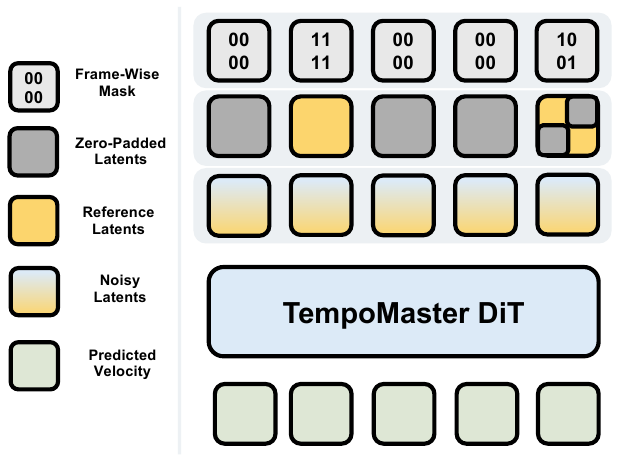}
   \caption{
   \textbf{Multi-Mask Condition.}
   Condition frames are zero-padded to the length of the full sequence; their latent representations and a frame-wise mask that provides precise timestep information are then concatenated with the noisy latents to guide generation.
   }
   \label{fig:4}
\end{figure}

\subsection{Multi-Mask Diffusion Transformer} \label{method.2}

The formulation of our method necessitates a model capable of processing dynamic conditions, including a single text prompt (with an image) for the initial stage and multiple frames (video) for the refinement stages. 

Condition injection via adapters or in-context learning generally requires dedicated training tailored to each specific condition, which presents a limitation to the flexibility and efficiency in both training and inference.
We therefore propose Multi-Mask, a unified framework for all aforementioned conditions. In contrast to other injection styles, our method avoids introducing extra parameters or increasing context length.
As illustrated in Fig.~\ref{fig:4}, Multi-Mask allows an arbitrary number of condition frames at any timestep. 
The condition frames maintain their temporal positions within a zero-padded sequence that matches the target video length. 
The padded sequence is then encoded into latent representations, which are concatenated with the noisy latents along the channel dimension. This design ensures temporal alignment at the latent level by construction. 
Additionally, a frame-wise mask is concatenated to provide fine-grained timestep information, mitigating the temporal ambiguity caused by the VAE's compression.
This versatile formulation allows a single model to handle multiple tasks—including T2V (Text-to-Video), I2V (Image-to-Video), FLF2V (First-Last-Frame-to-Video), and video continuation—as special cases of the Multi-Mask conditioning.

\subsection{Training} \label{method.3}

To improve generation quality and training stability, we utilize a two-stage training paradigm. 
In the first stage, the model learns to handle Multi-Mask conditioning from videos at a single frame rate. 
In the second stage, it learns to perform next-frame prediction from videos of multiple frame rates. 
Both stages are trained with the flow matching loss:
\begin{equation}
\label{eq:flow_matching}
\mathcal{L}_{\text{FM}}(\theta) = \mathbb{E}_{t, p_t(\mathbf{z}_0)} \left[ \lVert \mathbf{v}_\theta(\mathbf{z}_t, t) - (\mathbf{z}_1 - \mathbf{z}_0) \rVert_2^2 \right],
\end{equation}
where $\mathbf{z}_t = (1 - t) \mathbf{z}_0 + t \mathbf{z}_1$, with $t \sim \mathcal{U}[0, 1]$, $\mathbf{z}_0$ denoting clean video latents and $\mathbf{z}_1$ gaussian noise.

\paragraph{Single-Frame-Rate Training.} In this stage, the model is first trained to gain the ability to execute the complete denoising trajectory under the guidance of Multi-Mask conditions.  
For this training stage, we use short video clips of 121 frames at 24 fps.
From each clip, 0\% to 15\% of the frames are randomly selected and used as Multi-Mask conditions.

\paragraph{Multi-Frame-Rate Training.}
This stage equips the model with the capability to generate videos at variable frame rates. 
We use video clips with durations ranging from 5 seconds to 1 minute. 
For each clip, we randomly select a frame rate from 6, 12, 24 fps for training.
As shown in Fig.~\ref{fig:3}, we treat frame rate control as the manipulation of inter-frame intervals and inject this information via a modified Rotary Position Embedding (RoPE). Specifically, for a video sample $V^i$ with a target interval, we set the spacing between consecutive temporal position indices to $2^i$, thus aligning the positional index sequence with the video's timeline.
To ensure robustness, we employ a training-time augmentation that samples positional encodings from a broad, continuous range. This prevents the model from overfitting to a specific set of temporal indices and encourages it to learn a continuous positional function. As a result, the model generalizes effectively and demonstrates strong extrapolation capabilities in the temporal domain. Formally, the temporal position index for the $j$-th frame in $V^i$ is defined as
\begin{equation}
\begin{aligned}
&t_j = t_{\text{start}} + j \cdot 2^i,\, t_{\text{start}} \sim \mathcal{U}[0, T_{max}]
\end{aligned}
\end{equation}

\begin{figure}[t]
  \centering
   \includegraphics[width=\linewidth]{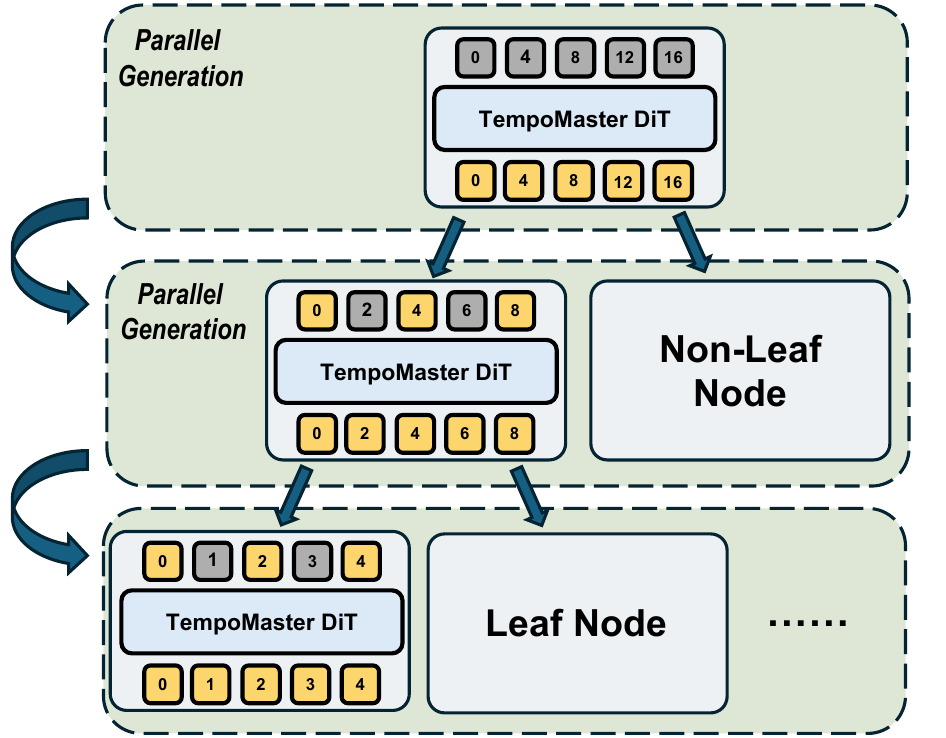}

   \caption{
   \textbf{The inference process of TempoMaster.} TempoMaster first generates videos with the lowest frame rate and the largest interval of temporal position indices. Within the same level, the generated frames can be partitioned into multiple segments to enable parallel generation, which proceeds hierarchically down to the leaf node level.
   }
   \label{fig:5}
\end{figure}

\paragraph{Training data.} 
The model is trained on a large, diverse in-house dataset comprising approximately 3 million high-quality clips, curated from a variety of web sources.
This dataset encompasses a broad spectrum of visual domains (e.g., indoor, outdoor, urban, natural) and clip durations (ranging from several to hundreds of seconds).

\begin{figure*}
    \centering
    \includegraphics[width=\textwidth]{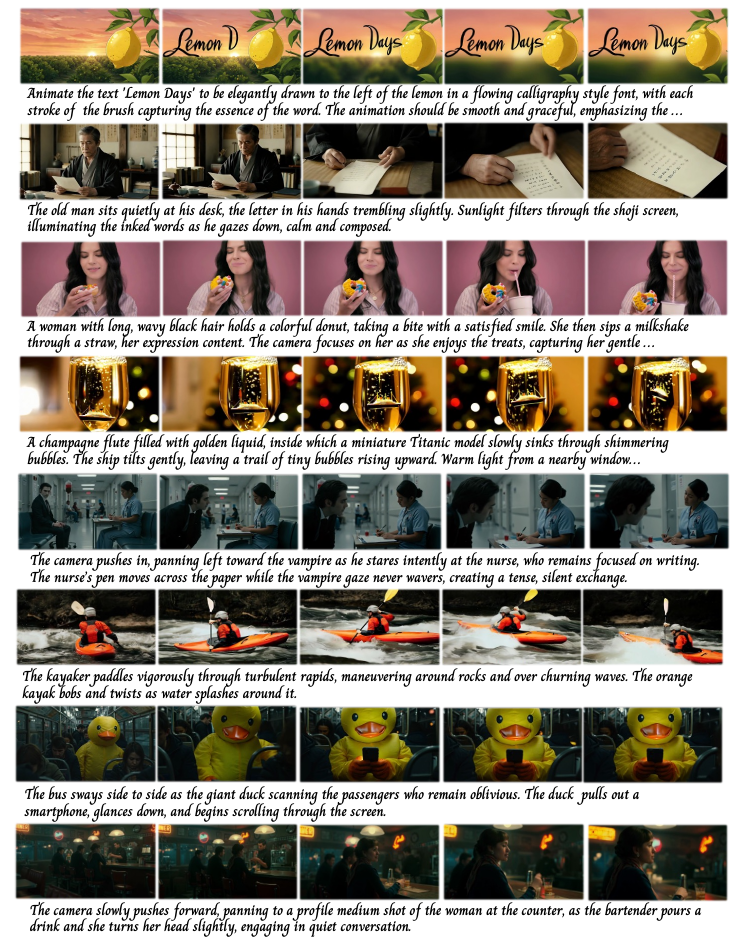}
    \caption{ 
    \textbf{Visualization of the generated videos.} 
    All videos are 500 frames in length and generated in 480p resolution. 
    }
    \label{fig:qualiative_results}
\end{figure*}

\begin{table*}[h]
    \centering
    \caption{
    \textbf{Vbench evaluation results.} We compare TempoMaster with state-of-the-art long video generation models of similar or larger size. Top: the evaluation results of long videos (500 frames). Bottom: the evaluation results of short videos (121 frames). Our method achieves the highest total score in both long and short video generation. 
    Higher values are better for all dimensions.
    }    
    \small
    \setlength{\tabcolsep}{8pt}  
    \begin{tabular}{l c| c c c c c c c}
        \toprule
        & & \multicolumn{7}{c}{\textbf{Vbench Evaluation} $\uparrow$}  \\

        \multicolumn{1}{l}{\multirow{-3}{*}{\textbf{Model}}}
        & \multicolumn{1}{l|}{\multirow{-3}{*}{\textbf{\#Params}}}
        & \shortstack{\textbf{Total} \\ \textbf{Score}} 
        & \shortstack{Subject \\ Consistency} 
        & \shortstack{Background \\ Consistency} 
        & \shortstack{Motion \\ Smoothness} 
        & \shortstack{Dynamic \\ Degree} 
        & \shortstack{Imaging \\ Quality} 
        & \shortstack{Aesthetic \\ Quality} \\
        
        \midrule
        MAGI-1 \cite{magi2025} & 24B & 78.50 & 98.26 & 97.29 & 99.41 & 21.38 & 66.36 & 55.91  \\
        FramePack \cite{framepack2025} & 13B & 79.52 & 98.68 & 99.20 & 99.54 & 16.82 & 70.90 & 61.34  \\
        SkyReels-V2 \cite{skyreelsv2_2025} & 14B & 79.17 & 96.04 & 96.01 & 99.07 & 53.28 & 64.85 & 56.28   \\
        MMPL \cite{mmpl2025} & 14B & 78.80 & 96.25 & 95.36 & 98.82 & 49.26 & 66.48 & 55.80  \\
        Ours & 14B & 80.30 & 97.41 & 97.87 & 98.94 & 41.10 & 70.20 & 59.62  \\ 
        \midrule
        MAGI-1 \cite{magi2025} & 24B & 79.05 & 97.66 & 97.63 & 99.28 & 32.24 & 67.71 & 58.49 \\
        FramePack \cite{framepack2025} & 13B & 79.90 & 98.24 & 99.03 & 99.49 & 22.87 & 70.77 & 61.48 \\
        SkyReels-V2 \cite{skyreelsv2_2025} & 14B &80.54 & 96.40 & 96.99 & 99.01 & 49.93 & 69.31 & 59.54 \\
        MMPL \cite{wan2025} & 14B & 80.55 & 95.67 & 95.10 & 98.60 & 69.25 & 67.24 & 56.78  \\
        Ours & 14B & 80.76 & 98.24 & 98.02 & 99.05 & 37.37 & 70.95 & 61.71  \\ 
        \bottomrule
    \end{tabular}
    
    \label{tab1:Vbench}
\end{table*}

\begin{figure*}
    \centering
    \includegraphics[width=\textwidth]{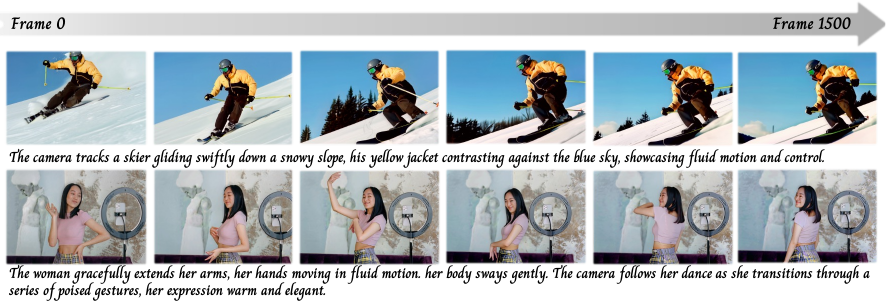}
    \caption{
    \textbf{Visualization of long-term generation stress test.} 
    Our method is capable of extending video clips with a window size comparable to autoregressive methods.  Our method composes minute-long videos (exceeding 1500 frames) by extending 480 frames with 5-second overlaps. 
    }
    \label{fig:long_video_test}
\end{figure*}

In typical data curation pipelines for large-scale video training, raw videos are processed by detecting shot transitions and subsequently segmented into single-shot clips \cite{wan2025, waver2025, moviegen2024}. While in real-world scenarios, videos longer than several seconds often comprise multiple shot transitions. 
Training solely on single-shot clips thus increases the difficulty of collecting long video data, substantially hindering scalability.
In our dataset, there are more than 300K videos curated from films, documentaries, and TV series, each with at least one shot transition. 

To ensure caption accuracy, we first segment the videos into single-shot clips and remove blurred frames from transitions.
We then captioned each clip in detail and concatenated its frames into new sequences to simulate cut transitions. 
To indicate temporal cuts, we connect captions with a \texttt{<Scene Cut>} token. 
Additionally, all multi-shot captions are prefixed with a \texttt{<MultiShot>} tag. 

However, during the training phase, we observed that directly employing multi-mask training on multi-shot videos produced unsatisfactory performance. 
We attribute this failure to information leakage: the random frame sampling in multi-mask training allows the model to observe all shots and thereby weakens the need to learn shot transitions.
Thus, for training on multi-shot videos, we randomly select shots and discard all conditioning frames within those shots. 
This augmentation forces the model to learn to generate new shots, thereby building an inherent capability for multi-shot video generation.

\subsection{Parallel Inference} \label{method.4}

Building on our temporal progressive generation framework, we introduce a parallel inference strategy to accelerate long video synthesis.
Fig.~\ref{fig:5} illustrates Tempo's generation procedure via a multiway tree. 
Each node maps to a time interval and its corresponding frames. 
Since the content of a time interval is predetermined by its parent node, child nodes within the same level exhibit no causal dependencies. 
Thus, they can be generated in parallel by dividing their parent node's frames into multiple segments.

Let $N$ denote all the frames to generate. Applying bidirectional attention on the full sequence results in a time complexity of $O(N)^2$.
Without loss of generality, we analyze a constrained variant of our parallel strategy in which the branching factor is uniform across all parent nodes. We define $W$ as the number of children per parent node, forming a $W$-way tree. The $i$-th level of this tree operates on a segment of $\frac{N}{2^{K-i} \cdot W^i}$ frames. Consequently, the overall computational complexity forms a geometric progression:
\begin{equation}
\begin{aligned}
\sum_{i=0}^{K-1} W^i \cdot \left( \frac{N}{2^{K-i} \cdot W^i} \right)^2 = \frac{N^2}{4^K} \cdot \sum_{i=0}^{K-1} \left( \frac{4}{W} \right)^i
\end{aligned}
\end{equation}

When $W \ge 4$, the sum of geometric progression becomes a constant, leading to an exponential acceleration with the overall complexity of $O\left( N^2/{4^K} \right)$.
Taking intra-level parallel generation into account, the overall complexity can be simply rewritten to:
\begin{equation}
\frac{N^2}{4^K} \cdot \sum_{i=0}^{K-1} \left( \frac{4}{W^2} \right)^i 
\end{equation}
This allows us to relax the requirement on $W$ to $W \ge 2$, which still ensures a geometric progression that converges to a constant. 
In our experiments, with $W$ set to 2, videos are generated in three sequential stages at 6, 12, and 24 fps by default.
The computational load of our method can be further reduced by omitting one or more intermediate refinement stages or by reducing the denoising steps within subsequent stages.
For instance, our model can generate $V^0$ directly from $V^{K-1}$. We analyze the performance and efficiency of these accelerated variants in Section~\ref{sec:expr}.

\section{Experiments}
\label{sec:expr}

\begin{table}
    \caption{
    \textbf{Human study results on long videos.}
    All videos are 500 frames in length and generated in 480p. Note that LongCat is a short-video generation model that can extend video clips with an overlap of only 13 frames. We employ it as a strong baseline for benchmarking aesthetic quality, semantic alignment, and motion quality. Each dimension is scored from 1 to 5 (with 1 representing the poorest quality and 5 the best), and the total score is computed as the average across all dimensions.
    }    
    \small
    \setlength{\tabcolsep}{2pt}  
    \begin{tabular}{l | c c c c c}
        \toprule
        & \multicolumn{5}{c}{\textbf{Human Study} $\uparrow$}  \\

        \multicolumn{1}{l|}{\multirow{-3}{*}{\textbf{Model}}}
        & \shortstack{Total \\ Score} 
        & \shortstack{Aesthetic \\ Quality} 
        & \shortstack{Semantic \\ Alignment} 
        & \shortstack{Motion \\ Quality} 
        & \shortstack{Content \\ Consistency} \\
        
        \midrule
        
        FramePack \cite{framepack2025} & 3.39 & \textbf{3.73} & 3.28 & 2.88 & 3.67 \\
        LongCat \cite{longcat2025} & 3.58 & 3.72 & 3.83 & 3.43 & 3.34 \\
        SkyReels-V2 \cite{skyreelsv2_2025} & 3.11 & 3.24 & 3.53 & 3.02 & 2.64  \\
        MMPL \cite{magi2025} & 2.93 & 3.19 & 3.43 & 2.65 & 2.44  \\
        Ours & \textbf{3.69} & 3.71 & \textbf{3.92} & \textbf{3.45} & \textbf{3.68}  \\ 
        \bottomrule
    \end{tabular}
    
    \label{tab2:human_study}
\end{table}

\begin{table*}[h]
    \centering
    \caption{
    \textbf{Ablation on the parallel configs.} Top: the evaluation results of long videos (500 frames). Bottom: the evaluation results of short videos (121 frames). We include total computational Flops for comparison. All videos maintain a resolution of 480p. Higher values are better for all dimensions.
    }    
    \small
    \setlength{\tabcolsep}{6pt}  
    \begin{tabular}{l c| c c c c c c c}
        \toprule
        & & \multicolumn{7}{c}{\textbf{Vbench Evaluation} $\uparrow$}  \\

        \multicolumn{1}{l}{\multirow{-3}{*}{\textbf{Method}}}
        & \multicolumn{1}{l|}{\multirow{-3}{*}{\textbf{PFLOPs}}}
        & \shortstack{\textbf{Total} \\ \textbf{Score}} 
        & \shortstack{Subject \\ Consistency} 
        & \shortstack{Background \\ Consistency} 
        & \shortstack{Motion \\ Smoothness} 
        & \shortstack{Dynamic \\ Degree} 
        & \shortstack{Imaging \\ Quality} 
        & \shortstack{Aesthetic \\ Quality} \\
        
        \midrule
        $f(6,24)m(1,4)$ & 526.48 & 80.30 & 97.91 & 97.57 & 98.74 & 43.97 & 69.49 & 59.35  \\
        $f(6,24)m(1,8)$ & 400.19 & 80.29 & 97.30 & 97.86 & 98.77 & 41.26 & 70.22 & 60.20   \\
        $f(6,12,24)m(1,2,4)$ & 774.23 & 80.30 & 97.41 & 97.87 & 98.94 & 41.10 & 70.20 & 59.62  \\
        $f(6,12,24)m(1,4,8)$ & 460.02 & 80.30 & 97.40 & 97.77 & 98.99 & 40.69 & 70.02 & 60.00  \\ 
        $f(6,12,24)m(1,8,8)$ & 539.57 & 80.04 & 97.38 & 97.84 & 99.00 & 40.61 & 70.00 & 59.96   \\
        \midrule
        $f(6,24)m(1,4)$ & 74.05 & 80.55 & 97.79 & 98.26 & 99.16 & 35.47 & 70.68 & 61.56  \\
        $f(6,24)m(1,8)$ & 66.91 & 80.46 & 97.80 & 98.25 & 99.13 & 35.27 & 70.59 & 61.39   \\
        $f(6,12,24)m(1,2,4)$ & 108.89 & 80.76 & 98.24 & 98.02 & 99.05 & 37.37 & 70.95 & 61.71  \\
        $f(6,12,24)m(1,4,8)$ & 96.99 & 80.26 & 97.82 & 98.26 & 99.26 & 32.42 & 70.50 & 61.33  \\ 
        $f(6,12,24)m(1,8,8)$ & 95.13 & 80.30 & 97.27 & 97.71 & 99.02 & 38.52 & 70.39 & 60.79  \\
        \bottomrule
    \end{tabular}
    
    \label{tab3:parallel_ablation}
\end{table*}

\begin{table}
\centering
    \caption{
    \textbf{Ablation on randomized temporal position indices.} We report Vbench metrics. All videos are 500 frames in length and maintain a resolution of 480p.
    }    
    \small
    \setlength{\tabcolsep}{3pt}
    \begin{tabular}{l | c c c c c c c}
        \toprule
        \textbf{Method} 
        & Tot. 
        & Subj.
        & Back.
        & Mot.
        & Dyn.
        & Imag.
        & Aes. \\
        
        \midrule
        w/o random & 80.00 & 97.41 & 97.87 & 98.94 & 37.70 & 70.20 & 59.62  \\
        w random & \textbf{80.19} & \textbf{97.44} & \textbf{97.94} & \textbf{98.96} & \textbf{39.09} & \textbf{70.23} & \textbf{59.74} \\
        \bottomrule
    \end{tabular}
    
    \label{tab4:random_timestep_ablation}
\end{table}

\subsection{Implementation Details}
\paragraph{Set Up.} We implement TempoMaster based on Wan2.2~\cite{wan2025}, a Mixture-of-Experts (MoE) architecture that separates the denoising process across timesteps with two specialized trained Diffusion Transformers. 
We choose the model trained on high noise as our base model for simplicity, which takes an image as input and denoises at an extremely high noise level. 
As mentioned in Sec.~\ref{method.3}, the training consists of two stages. The single-temporal-resolution training stage takes about 300 H100 GPU days, and the multi-temporal-resolution stage takes another 1200 H100 GPU days. 
Both stages are optimized using the AdamW optimizer, with weight decay of 1e-4 and learning rates of 5e-4 and 2e-5, respectively.
All experiments are performed on the Teletron framework~\cite{TeleTron2025}.

\paragraph{Evaluation Metrics.} We comprehensively evaluate the generation capability of our model on both long videos (500 frames) and short videos (121 frames). We choose Vbench-Long~\cite{Vbench2023, Vbench++2024} quality metrics for automatic evaluation, including subject consistency, background consistency, motion smoothness, dynamic degree, imaging quality, and aesthetic quality. The total score is calculated using the same numeric system as Vbench. To ensure a fair and reproducible comparison, all videos are generated under the standard Vbench I2V test suite conditions, with a uniform 480p resolution and 16:9 aspect ratio. We further curate a set of 150 image-text pairs from real-world users and professional artists to further assess performance in complex, realistic scenarios through a human study. 20 participants are involved to evaluate the generated videos based on human perception, especially focusing on aesthetic quality, semantic alignment, motion quality, and content consistency.

\subsection{Comparing with State-of-the-arts}
We compare our model with existing state-of-the-art long video generation models of similar size, including MAGI-1-24B, FramePack, Skyreels-V2, and MMPL. 
As shown in Table \ref{tab1:Vbench}, TempoMaster achieves the highest total score in both long video and short video generation. 
It is noteworthy that autoregressive-like methods (e.g., MAGI, SkyReels-V2) exhibit a significant performance drop in long-video evaluation compared to their short-video metrics, which can be attributed to the temporal drifting caused by accumulating errors during iterative generation.
However, Vbench metrics such as subject consistency, background consistency, and motion smoothness consistently favor videos with limited motion.
Gaining a higher score on this evaluation does not necessarily correspond to superior overall video quality. 
We therefore use a human study as our primary benchmark for perceptual evaluation. As shown in Tab.~\ref{tab2:human_study}, TempoMaster achieves a higher overall score than all strong baselines, with notable advantages in semantic alignment, motion quality, and content consistency. Note that methods like FramePack perform significantly worse on motion quality in human evaluation compared to Vbench, due to the limited dynamics in their outputs.

\subsection{Qualitative Experiments}

We present qualitative results to illustrate the generation quality of TempoMaster in this section. 
Fig.~\ref{fig:qualiative_results} qualitatively demonstrates our model's ability to generate long, high-quality videos, including strong performance in motion dynamics, visual detail, and prompt adherence.
To further evaluate the model's capacity for generating extremely long video sequences, we employ a video continuation strategy under the Multi-Mask framework. 
This approach utilizes the last frames of each preceding segment as initial conditions for the subsequent segment, enabling seamless long-term generation. 
For each continuation step, the model takes the last 5 seconds of the previous segment as input and generates 480 frames.
As shown in Fig.~\ref{fig:long_video_test}, our model maintains temporal coherence and visual stability over 1500 generated frames without significant degradation.

\subsection{Ablation Study}

\paragraph{Ablation on parallel strategy on inference.}
As outlined in Sec.~\ref{method.4}, the inference process comprises $K$ stages corresponding to videos with $K$ frame rates, and the generation of the $i$-th stage can be partitioned into $M_i$ segments for parallel computation.
We define each unique parallel configuration by a frame rate list of length $K$ and its corresponding set of parallelism factors $M_{i=1}^K$. For example, the configuration $f(6,12,24)m(1,2,4)$ denotes the inference consists of 3 stages, where the frame rates are $[6,12,24]$ and the parallel segments for each stage are $M_1=1$, $M_2=2$, and $M_3=4$, respectively.
We report the Vbench metrics and total computational PFLOPs ($10^{15}$ floating-point operations) under different parallel configurations in Tab.~\ref{tab3:parallel_ablation}. The performance is robust to different parallel configurations, demonstrating that inference can be accelerated without qualitative degradation.

\paragraph{Ablation on Randomized Temporal Position Indices.}

We conduct an ablation study on our randomized temporal index strategy. As reported in Tab.~\ref{tab4:random_timestep_ablation}, the method with randomization consistently outperforms the baseline across Vbench metrics with the same training budget.

\section{Conclusion}
\label{sec:conc}
In this paper, we propose TempoMaster, a novel framework that reformulates long video generation through a next-frame-rate prediction paradigm. By decoupling global content planning from local detail refinement, our method effectively mitigates the issue of error accumulation inherent in conventional next-frame prediction models. Generating global contents first not only brings higher temporal consistency but also unlocks parallel generation in the subsequent refining stages. Experimental results demonstrate that TempoMaster achieves state-of-the-art performance in generating long videos with rich dynamics and superior temporal consistency, offering a practical and strong solution for real-world long video generation.

Limitations \& Future Work.
The current framework leverages parallel generation to accelerate inference but does not specifically optimize Time-To-First-Frame (TTF) latency. This limitation may restrict its applicability in real-time video streaming scenarios. Future work will focus on reducing TTF through streaming generation and adaptive scheduling.
\appendix
\appendix
\newpage
\section*{Appendix}
\begin{table}[]
    \centering
    \small
    \setlength{\tabcolsep}{3pt}  
    \begin{tabular}{c|c c c c c}
    \toprule
    \shortstack{Training \\ Stage} & \shortstack{Training \\ Steps} & \shortstack{Batch \\ Size} & \shortstack{Warmup \\ Steps} & \shortstack{Weight \\ Decay} & \shortstack{EMA \\ Weight}
    \\
    \midrule
    Single-Frame-Rate & 15000 & 32 & 2000 & 1e-4 & 0.999 \\
    Multi-Frame-Rate & 45000 & 32 & 2000 & 1e-4 & 0.999 
    \\
    \bottomrule
    \end{tabular}
    \caption{\textbf{Additional hyperparameters in training.} We enable ema for a more consistent training.}
    \label{tab:training_hyperparameters}
\end{table}

\section{Training Details}

\paragraph{Base Model}
We employ the high-noise model of Wan2.2~\cite{wan2025} as our base model, which can only denoise at an extremely high noise level. 
In contrast to its predecessor Wan2.1, this model eliminates the dependency on CLIP~\cite{clip2021} features extracted from the first frame. This structural simplicity makes it more suitable for our Multi-Mask conditioning framework.
Since the base model is inherently limited to extreme noise levels, we first adapt it through a training phase that enables denoising across all noise levels, as mentioned in this paper.

\paragraph{Hyperparameters}
We employ the AdamW optimizer with a consistent learning rate schedule, using learning rates of 5e-4 and 2e-5 for the respective training stages.
We further provide additional hyperparameters during training in Tab.~\ref{tab:training_hyperparameters}. 

\paragraph{Noise Scheduling}
During training, we adopt the logit-normal distribution over $t$ following~\cite{sd2024}: 
\begin{equation}
    \pi_{\text{ln}}(t;m,s)=\dfrac{1}{s\sqrt{2\pi}}\dfrac{1}{t(1-t)}\exp{-\dfrac{(\text{logit}(t) - m)^2}{2s^2}}
\end{equation}
Specifically, we set $m$ to 0 and $s$ to 1, and employ a sigma shift of 3 during training.

\begin{figure*}
    \centering
    \includegraphics[width=0.9\textwidth]{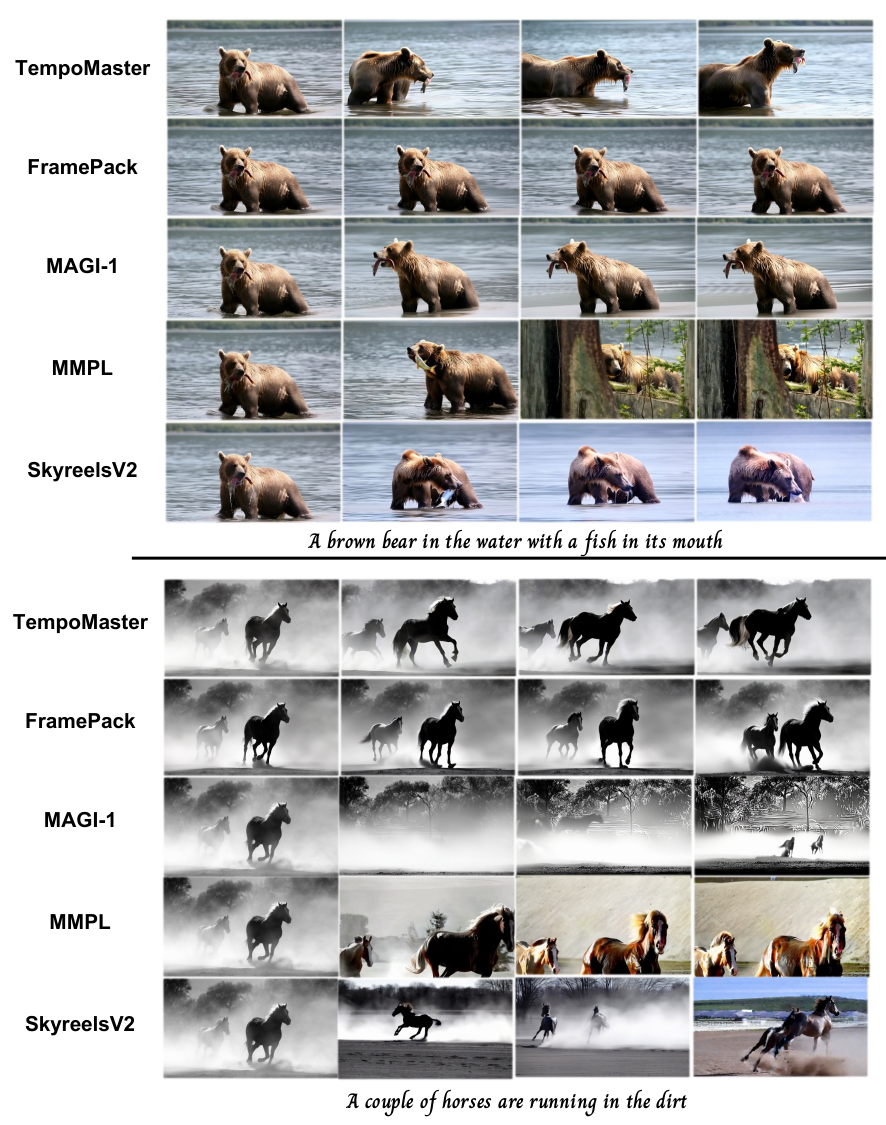}
    \caption{
    \textbf{Qualitative comparisons.} 
    We compare TempoMaster with representative long video methods. 
    }
    \label{fig:comparison}
\end{figure*}

\section{Training Data}
Our data pipeline draws upon existing short video methods~\cite{moviegen2024, seedance2025, wan2025, waver2025}.
To ensure the high quality of training data, we filter videos through multi-dimensional quantitative evaluation:

\begin{itemize}
\item \textbf{Aesthetic Assessment}: Video frames are evaluated through the average score from an image aesthetic model. This process guarantees that the training data possesses high aesthetic value, thereby enhancing the visual appeal of the generated content.
\item \textbf{Clarity Detection}: This module primarily employs the Laplacian operator to quantify image sharpness. It focuses on monitoring detail fidelity in dynamic scenes, effectively identifying and filtering out blurred sequences to maintain high image quality across the training samples.
\item \textbf{Motion Analysis}: The coherence and magnitude of motion are assessed by computing optical flow between consecutive frames. This allows for the effective exclusion of static frames and motion-distorted content, thereby improving the dynamic expressiveness of the dataset.
\end{itemize}

\section{The Principle of Human Study}

The human evaluation study is designed to assess the quality of generated long videos across multiple key dimensions, reflecting both aesthetic and functional performance. Each video is rated on a scale from 1 to 5 across the following four dimensions:

\begin{itemize}
\item \textbf{Aesthetic Quality}: Evaluates visual appeal through composition, clarity, lighting, and detail rendering. Penalties apply for overexposure, clutter, artifacts, or blurriness. Severe flaws receive the lowest score.

\item \textbf{Semantic Alignment}: Measures fidelity to the text prompt, including background, action, and lighting. 
Minor deviations result in a 1-point deduction, while partial or complete failure to execute the described actions results in scores of 1–2.

\item \textbf{Motion Quality}: Assesses movement amplitude, speed, and plausibility. Deductions occur for static frames, incoherent motion, implausible dynamics, or severe artifacts.

\item \textbf{Content Consistency}: Tracks temporal degradation in subject appearance, motion degradation (e.g., incoherence or repetition), and visual decay (e.g., color shift or blurring). Significant drift or decay leads to a score of 1.
\end{itemize}

\section{Additional Visualization Results}

We provide more qualitative comparison results with prior works in Fig.~\ref{fig:comparison}. We observe that FramePack outputs are often characterized by a lack of dynamic motion, whereas the other compared methods exhibit marked degradation in visual quality.
{
    \small
    \bibliographystyle{ieeenat_fullname}
    \bibliography{main}
}

\end{document}